\documentclass[acmlarge,nonacm,screen]{acmart}
\usepackage{listings}
\usepackage{xcolor}
\usepackage{tikz}
\usepackage{pgfplots}
\pgfplotsset{compat=1.18}
\usetikzlibrary{shapes.geometric, arrows.meta, positioning, calc, backgrounds, fit, decorations.pathreplacing}

\definecolor{layerstructural}{HTML}{4A7C9B}
\definecolor{layercontent}{HTML}{D4883A}
\definecolor{layerrouting}{HTML}{5B8A72}
\definecolor{reviewgate}{HTML}{C45B4A}
\definecolor{stagebox}{HTML}{4A7C9B}
\definecolor{lightgray}{HTML}{F5F5F5}



\title{Interpretable Context Methodology: Folder Structure as Agent Architecture}

\author{Jake Van Clief, David McDermott}
\email{theceo@eduba.io}
\affiliation{%
  \institution{Eduba, University of Edinburgh}
  \city{Palm Coast}
  \state{Florida}
  \country{USA}
}

\begin{document}

\begin{abstract}
Current approaches to AI agent orchestration typically involve building multi-agent frameworks that manage context passing, memory, error handling, and step coordination through code. These frameworks work well for complex, concurrent systems. But for sequential workflows where a human reviews output at each step, they introduce engineering overhead that the problem does not require. This paper presents Interpretable Context Methodology (ICM), a method that replaces framework-level orchestration with filesystem structure. Numbered folders represent stages. Plain markdown files carry the prompts and context that tell a single AI agent what role to play at each step. Local scripts handle the mechanical work that does not need AI at all. The result is a system where one agent, reading the right files at the right moment, does the work that would otherwise require a multi-agent framework. This approach applies ideas from Unix pipeline design, modular decomposition, multi-pass compilation, and literate programming to the specific problem of structuring context for AI agents. The protocol is open source under the MIT license.\footnote{\url{https://github.com/RinDig/Interpretable-Context-Methodology-ICM-}}
\end{abstract}

\maketitle

\section{Introduction}

There are genuinely good agentic frameworks available today. CrewAI, LangChain, AutoGen, and others handle multi-step orchestration, memory management, tool use, and error recovery. They work. But they work within their own structures, and adjusting those structures requires development work. Changing the order of steps, swapping a prompt, adding or removing a stage, skipping something that is not relevant today: these actions typically mean editing code, understanding abstractions, and redeploying. For practitioners whose workflows are sequential and need human review at each step, the control surface can be much simpler.

This paper describes Interpretable Context Methodology (ICM), a method for orchestrating AI agent workflows using folder structure, markdown files, and local scripts. The central observation is straightforward: if the prompts and context for each stage of a workflow already exist as files in a well-organized folder hierarchy, you do not need a coordination framework to manage multiple specialized agents. You need one orchestrating agent that reads the right files at the right moment. The folder structure tells it what to do at each step, and if the agent delegates sub-tasks, the same folder structure determines what context those sub-agents receive. Local Python scripts handle the parts that do not need AI: fetching data, moving files, formatting output, sending emails.

This is going backward before going forward. The principles that made Unix pipelines effective in the 1970s\footnote{Programs that do one thing. Output of one becomes input of another. Plain text as universal interface. These ideas are over fifty years old and they hold up.} and multi-pass compilers tractable in the 1980s apply directly to AI agent orchestration in the 2020s. ICM applies those principles to the specific challenge of structuring context for language models.

The central question this paper examines is how structuring the context delivery mechanism as a filesystem hierarchy affects practitioners' ability to control, inspect, and edit AI agent behavior across multi-step workflows, and what this structure means for the quality of the model's output at each stage.

The paper is organized as follows. Section 2 traces the relevant background across software engineering, context engineering, and human oversight research. Section 3 describes the protocol itself. Section 4 walks through working implementations and reports on early practitioner experience. Section 5 discusses where this approach fits and where it does not, including implications for the design of interactive intelligent systems more broadly. Section 6 explores future directions, drawing on the structural parallels between ICM and multi-pass compilation to propose semantic debugging and source-level traceability for AI workflows.

\begin{table}[t]
\centering
\caption{Comparison of control surfaces for sequential, human-reviewed workflows. The first six rows show dimensions where ICM's filesystem approach simplifies common operations. The last four rows show dimensions where framework-based approaches provide capabilities that ICM lacks or handles less well.}
\label{tab:comparison}
\begin{tabular}{p{3.2cm} p{4.5cm} p{4.5cm}}
\toprule
\textbf{Dimension} & \textbf{Framework approach} & \textbf{ICM approach} \\
\midrule
Change stage order & Edit orchestration code, redeploy & Rename or reorder folders \\
\addlinespace
Modify a prompt & Edit agent configuration in code & Edit a markdown file \\
\addlinespace
Add or remove a stage & Write new agent class, update orchestrator & Add or delete a folder \\
\addlinespace
Inspect intermediate state & Add logging, build dashboard & Open the folder, read the files \\
\addlinespace
Hand off to another person & Document environment, dependencies, setup & Copy the folder \\
\addlinespace
Who can make changes & Developer & Anyone with a text editor \\
\midrule
Error recovery mid-pipeline & Built-in retry, fallback, exception handling & Manual re-run of failed stage \\
\addlinespace
Conditional branching & Programmatic routing based on agent output & Human decides between stages \\
\addlinespace
Concurrent execution & Native parallel agent coordination & Sequential by design \\
\addlinespace
External service integration & Programmatic API calls, auth management & Local scripts or MCP connections \\
\bottomrule
\end{tabular}
\end{table}

\section{Background and Related Work}

\subsection{Composability and the Unix Tradition}

In 1978, Doug McIlroy articulated the principles that would define Unix's design philosophy: make each program do one thing well, expect the output of every program to become the input to another, and use text streams as the universal interface between programs \cite{mcilroy1978}. These principles were not theoretical. They were engineering decisions driven by constraints. The PDP-11 machines that ran early Unix had limited memory. Programs had to be small. The way to build powerful systems from small programs was to connect them through a common interface \cite{ritchie1974}.

Kernighan and Pike later argued that the power of a Unix system comes more from the relationships among programs than from the programs themselves \cite{kernighan1984}. Eric Raymond codified this into explicit design rules: the Rule of Modularity (write simple parts connected by clean interfaces), the Rule of Transparency (design for visibility to make inspection and debugging easier), and the Rule of Composition (design programs to be connected to other programs) \cite{raymond2003}.

These principles were formalized in software architecture as the ``pipe-and-filter'' pattern by Shaw and Garlan \cite{shaw1996}: a system of independent components, each reading from inputs and writing to outputs, connected by data streams. The pattern's strength is that any component can be replaced, inspected, or tested independently.

A related lineage runs through build systems. Stuart Feldman's Make (1979) established that workflows could be defined as dependency graphs between files using declarative specifications \cite{feldman1979}. The key insight: files are both the artifacts of work and the coordination mechanism between stages. You do not need a separate orchestration layer when the filesystem tracks what has been produced and what depends on it. Multi-pass compilers work on the same principle: source code transforms through a sequence of intermediate representations, each pass reading the output of the previous pass, with well-defined interfaces between them \cite{aho2006}.

David Parnas argued in 1972 that systems should be decomposed based on what each module hides from the rest of the system, yielding components that can be modified independently \cite{parnas1972}. Edsger Dijkstra coined the term ``separation of concerns'' to describe the discipline of addressing one thing at a time as the only available technique for effective ordering of one's thoughts \cite{dijkstra1974}.

These ideas appear across decades and contexts because they describe something real about how systems stay manageable as they grow. They are relevant here because the problem of orchestrating AI agents through multi-step workflows is, at its core, a problem of modular decomposition, clean interfaces, and readable intermediate representations.

\subsection{Context Engineering and Agentic AI}

The practitioner community has increasingly adopted the term ``context engineering'' to describe what building production AI systems actually involves. Andrej Karpathy gave the term its clearest articulation in June 2025, arguing that ``prompt engineering'' understates the work \cite{karpathy2025}. The distinction is useful. Prompt engineering suggests crafting a single instruction. Context engineering describes the broader discipline of filling the context window with the right information: instructions, retrieved knowledge, memory, tool descriptions, and prior outputs, all structured so the model can use them effectively. This paper uses the term in that sense.

Lance Martin at LangChain formalized this into a taxonomy of strategies: write (author instructions), select (choose relevant context), compress (reduce token waste), and isolate (keep unrelated context separate) \cite{martin2025}. Simon Willison argued that the entire information environment, including previous model responses and system state, is part of the context that needs engineering \cite{willison2025}.

The current generation of agentic frameworks, LangChain \cite{chase2022}, AutoGen \cite{wu2023autogen}, CrewAI, and others, handle context engineering through code-level abstractions. They define agents as objects, conversations as message arrays, and orchestration as programmatic control flow. This works well for systems that need dynamic multi-agent collaboration, concurrent execution, or complex branching logic.

But for sequential workflows, these frameworks solve a coordination problem that may not need to exist. If Agent A's job is to research, Agent B's job is to filter, and Agent C's job is to write, the framework's role is to pass the right context to the right agent at the right time. That coordination can also be achieved by putting the right files in the right folders. The orchestrating agent reads different instructions at each stage. If it delegates sub-tasks to smaller models (as current agent-team architectures allow), the folder structure provides the context for those delegations too. The coordination logic lives in the filesystem, not in application code.

This matters because of how language models handle context. Liu et al. demonstrated that LLMs perform significantly worse when relevant information is buried in the middle of long contexts \cite{liu2024}. The more irrelevant material in the context window, the worse the model performs on the material that matters. Jiang et al. showed that prompt compression can achieve up to 20x token reduction with minimal performance loss \cite{jiang2023}, but a simpler approach is to avoid loading irrelevant context in the first place. Stage-specific context loading, where each stage only sees the files it needs, prevents the problem rather than treating it after the fact.

It is worth distinguishing ICM from Anthropic's Model Context Protocol (MCP) \cite{anthropic2024mcp}. MCP standardizes how models access external tools and data sources, solving the integration problem between AI systems and the services they need to call. ICM addresses a different layer: how to structure and deliver context to an agent across a multi-stage workflow. The two are complementary. An ICM stage might use MCP connections to access external services, while the stage's folder structure determines what context the agent receives when doing so. This separation matters for efficiency as well. Jones and Kelly at Anthropic observed that loading all tool definitions upfront into the context window slows agents and increases costs \cite{jones2025}. ICM's stage-based architecture avoids this by scoping tool definitions to individual stages, loading only the tools relevant to the current step.

\subsection{Human Oversight and Observability}

The question of how humans should relate to automated systems has been studied for decades, and the findings are remarkably consistent.

Fails and Olsen introduced the interactive machine learning paradigm in 2003: rapid cycles of system output, human feedback, and correction \cite{fails2003}. Amershi et al. argued that interactive ML must involve users at all stages, from training through evaluation, with interfaces that support steering and correction \cite{amershi2014}. Dudley and Kristensson's review of interface design for interactive ML emphasized that transparent, inspectable representations are essential for effective human-AI collaboration \cite{dudley2018}.

Eric Horvitz's work on mixed-initiative systems established principles for coupling automated services with human control \cite{horvitz1999}. The key insight: systems should let users invoke, adjust, and terminate automated processes at natural breakpoints. This requires that the system's state be visible and its actions be reversible.

Parasuraman and Riley identified the failure modes that emerge when this goes wrong \cite{parasuraman1997}. When automated outputs are opaque, people either trust them blindly (misuse) or stop using them entirely (disuse). Both failures stem from the same cause: the human cannot see what happened between input and output. Lee and See's work on trust calibration reinforced this: appropriate trust requires that system behavior be observable \cite{lee2004}. Parasuraman, Sheridan, and Wickens proposed a taxonomy of automation levels, noting that the right level of automation varies by task and that systems should support different levels at different stages \cite{parasuraman2000}.

Ben Shneiderman synthesized these threads into the Human-Centered AI framework, arguing that systems can achieve both high human control and high automation simultaneously \cite{shneiderman2020}. The two are not in tension. They reinforce each other when the system is designed to be comprehensible, predictable, and controllable \cite{shneiderman2022}.

Cynthia Rudin made the most forceful version of this argument: stop building opaque systems and then trying to explain them after the fact. Build systems that are inherently interpretable \cite{rudin2019}. This applies at the workflow level as much as at the model level. A production pipeline where every intermediate output is a readable file is inherently interpretable. There is nothing to explain because nothing was hidden.

This is also becoming a regulatory concern. The EU AI Act requires human oversight of high-risk AI systems, distinguishing between human-in-the-loop, human-on-the-loop, and human-in-command approaches \cite{enqvist2023}. Novelli et al. argue that effective oversight requires institutional design, not just technical capability \cite{novelli2024}. Systems with staged review points, audit trails, and defined intervention surfaces have a practical advantage as these requirements take effect.

\section{Interpretable Context Methodology}

\subsection{Design Principles}

ICM is built on five principles, each borrowed from established practice.

\textbf{One stage, one job.} Each stage in a workspace handles a single step of the workflow and writes its output to its own folder. This follows McIlroy's Unix principle and Parnas's information-hiding criterion \cite{mcilroy1978, parnas1972}. A stage that fetches data does not also filter it. A stage that filters does not also format the final output. Each stage reads a defined input, transforms it, and writes a defined output, the same structure that governs individual passes in a multi-pass compiler.

\textbf{Plain text as the interface.} Stages communicate through markdown and JSON files. No binary formats, no database connections, no proprietary serialization. This follows Kernighan and Pike's argument that text is the universal interface \cite{kernighan1984}. Any tool that can read a text file can participate in the workflow. Any human who can open a text editor can inspect or modify any artifact.

\textbf{Layered context loading.} Agents load only the context they need for the current stage, following the principle that less irrelevant context means better model performance \cite{liu2024}. This is prevention rather than compression \cite{jiang2023}. Within the content layers, ICM further distinguishes between reference material (stable rules and conventions that persist across runs) and working artifacts (per-run content that changes every time). The model receives these as structurally separate context, which matters because they require different kinds of attention: reference material should be internalized as constraints, while working artifacts should be processed as input.

\textbf{Every output is an edit surface.} The intermediate output of each stage is a file a human can open, read, edit, and save before the next stage runs. This implements Horvitz's mixed-initiative principles \cite{horvitz1999} and Shneiderman's direct manipulation paradigm \cite{shneiderman1983}: the human works with visible, manipulable objects, and the system picks up whatever the human left there.

\textbf{Configure the factory, not the product.} A workspace is set up once with the user's preferences, brand, style, and structural decisions. After that, each run of the pipeline produces a new deliverable using the same configuration. This follows the continuous delivery principle that production pipelines should be repeatable \cite{humble2010}.

\subsection{Architecture}

\begin{figure}[t]
\centering
\begin{tikzpicture}[scale=0.82, every node/.append style={transform shape},
    layerbox/.style={
        minimum height=0.9cm,
        text=white,
        font=\sffamily\small,
        rounded corners=2pt,
        anchor=west,
        align=center
    },
    labeltext/.style={
        font=\sffamily\footnotesize,
        anchor=west,
        text=black
    },
    tokentext/.style={
        font=\sffamily\footnotesize,
        anchor=east,
        text=black!60
    }
]

% Layer boxes with progressive widths
\node[layerbox, fill=layerstructural, minimum width=5cm] (L0) at (0,0) {\textbf{Layer 0:} CLAUDE.md};
\node[layerbox, fill=layerstructural!85, minimum width=6.5cm] (L1) at (-0.75,-1.15) {\textbf{Layer 1:} CONTEXT.md};
\node[layerbox, fill=layerstructural!70, minimum width=8cm] (L2) at (-1.5,-2.3) {\textbf{Layer 2:} Stage CONTEXT.md};
\node[layerbox, fill=layercontent, minimum width=9.5cm] (L3) at (-2.25,-3.45) {\textbf{Layer 3:} Reference material};
\node[layerbox, fill=layercontent!80, minimum width=11cm] (L4) at (-3,-4.6) {\textbf{Layer 4:} Working artifacts};

% Right-side questions
\node[labeltext] at (5.2,0) {``Where am I?''};
\node[labeltext] at (5.95,-1.15) {``Where do I go?''};
\node[labeltext] at (6.7,-2.3) {``What do I do?''};
\node[labeltext] at (7.45,-3.45) {``What rules apply?''};
\node[labeltext] at (8.2,-4.6) {``What am I working with?''};

% Left-side token counts
\node[tokentext] at (-0.15,0) {\texttildelow800 tok};
\node[tokentext] at (-0.9,-1.15) {\texttildelow300 tok};
\node[tokentext] at (-1.65,-2.3) {200--500 tok};
\node[tokentext] at (-2.4,-3.45) {500--2k tok};
\node[tokentext] at (-3.15,-4.6) {varies};

% Bracket labels for structural vs content
\draw[decorate, decoration={brace, amplitude=6pt, mirror}, thick, layerstructural]
    (-3.3, 0.45) -- (-3.3, -2.75) node[midway, left=10pt, font=\sffamily\footnotesize, text=layerstructural, align=right] {Structural\\(routing)};
\draw[decorate, decoration={brace, amplitude=6pt, mirror}, thick, layercontent]
    (-3.3, -2.95) -- (-3.3, -5.05) node[midway, left=10pt, font=\sffamily\footnotesize, text=layercontent, align=right] {Content\\(factory / product)};

\end{tikzpicture}
\caption{The five-layer context hierarchy. Layers 0--2 provide structural routing and stage instructions. Layers 3 and 4 carry content: Layer~3 holds reference material (the factory), stable across runs; Layer~4 holds working artifacts (the product), unique to each run.}
\label{fig:layers}
\end{figure}
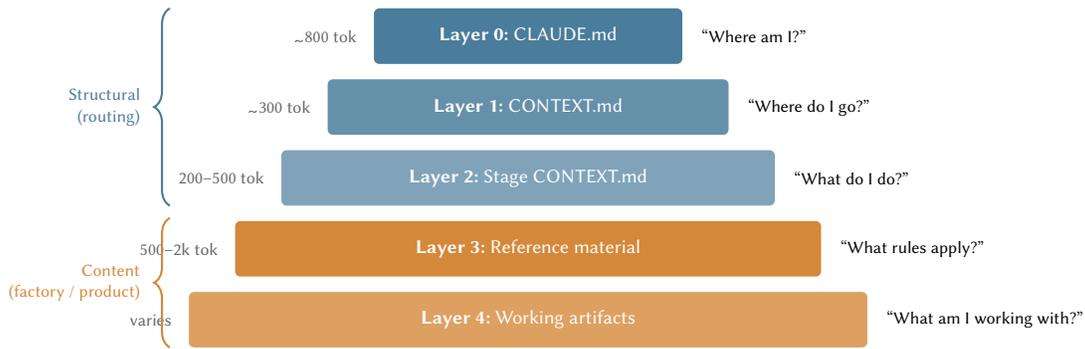

An ICM workspace is a folder. Inside it, agents navigate a five-layer context hierarchy (Figure~\ref{fig:layers}).

Layer 0 is the global identity file. It tells the agent which workspace it is in, what the folder structure contains, and where to find things. Layer 1 is workspace-level task routing: given what the user wants to do, which stage handles it, and what shared resources exist across stages. Layer 2 is stage-specific: the contract that defines inputs, process, and outputs for one step of the workflow.

Layers 3 and 4 are both content that the agent loads while executing a stage, but they represent fundamentally different kinds of context.

Layer 3 is reference material: design systems, voice rules, build conventions, style guides, domain knowledge bundled as skill files. These files are configured once during workspace setup and remain stable across every run of the pipeline. They are the factory.\footnote{This connects to the fifth design principle: configure the factory, not the product. Layer 3 is the factory configuration. Layer 4 is what the factory produces each time it runs.} Layer 4 is working artifacts: the output of the previous stage, user-provided source material, anything specific to this particular run of the pipeline. These files are produced and consumed during execution and change every time.

The distinction matters for how the model processes context. Layer 3 material needs to be internalized as constraints and patterns: the model should write \textit{like this}, use \textit{these colors}, follow \textit{these conventions}. Layer 4 material needs to be processed as input: the model should transform \textit{this research} into a script, or convert \textit{this script} into a visual specification. Mixing persistent rules with per-run artifacts in an undifferentiated context window forces the model to sort them on its own. Separating them in the folder structure means the model receives already-organized context.

\begin{table}[t]
\centering
\caption{Layer 3 (reference material) versus Layer 4 (working artifacts).}
\label{tab:layers34}
\begin{tabular}{p{3.5cm} p{4.8cm} p{4.8cm}}
\toprule
& \textbf{Layer 3: Reference} & \textbf{Layer 4: Working} \\
\midrule
Changes between runs & No & Yes \\
\addlinespace
Example files & voice.md, design-system.md, conventions.md & research-output.md, script-draft.md \\
\addlinespace
Model should & Internalize as constraints & Process as input \\
\addlinespace
Configured during & Workspace setup (once) & Pipeline execution (each run) \\
\addlinespace
Folder location & \texttt{references/}, \texttt{\_config/}, \texttt{shared/} & \texttt{output/} \\
\addlinespace
Analogy & The recipe & The ingredients \\
\bottomrule
\end{tabular}
\end{table}

A rendering agent might only need Layers 0 through 2. A script-writing agent reads down to Layer 4 to access both the voice rules (Layer 3) and the source material (Layer 4). No agent reads everything. This keeps token cost low and context focused, and it avoids the degradation that Liu et al.\ documented when models process long contexts full of irrelevant material \cite{liu2024}.

\begin{figure}[t]
\centering
\begin{tikzpicture}[
    every node/.style={font=\ttfamily\small, anchor=west},
    folder/.style={font=\ttfamily\small\bfseries, anchor=west},
    label/.style={font=\sffamily\footnotesize, text=black!50, anchor=west},
    layerlabel/.style={font=\sffamily\footnotesize\bfseries, rounded corners=2pt, inner sep=2pt, minimum height=0.4cm}
]

\def\ind{0.45}
\def\lh{0.5}

% Tree structure
\node[folder] (root) at (0,0) {workspace/};
\node (claude) at (\ind, -\lh) {CLAUDE.md};
\node (ctx1) at (\ind, -2*\lh) {CONTEXT.md};

\node[folder] (stages) at (\ind, -3*\lh) {stages/};
\node[folder] (s01) at (2*\ind, -4*\lh) {01\_research/};
\node (ctx2a) at (3*\ind, -5*\lh) {CONTEXT.md};
\node (ref1) at (3*\ind, -6*\lh) {references/};
\node (out1) at (3*\ind, -7*\lh) {output/};

\node[folder] (s02) at (2*\ind, -8*\lh) {02\_script/};
\node (ctx2b) at (3*\ind, -9*\lh) {CONTEXT.md};
\node (ref2) at (3*\ind, -10*\lh) {references/};
\node (out2) at (3*\ind, -11*\lh) {output/};

\node[folder] (s03) at (2*\ind, -12*\lh) {03\_production/};
\node (ctx2c) at (3*\ind, -13*\lh) {CONTEXT.md};
\node (ref3) at (3*\ind, -14*\lh) {references/};
\node (out3) at (3*\ind, -15*\lh) {output/};

\node[folder] (config) at (\ind, -16*\lh) {\_config/};
\node[folder] (shared) at (\ind, -17*\lh) {shared/};
\node[folder] (setup) at (\ind, -18*\lh) {setup/};
\node (quest) at (2*\ind, -19*\lh) {questionnaire.md};

% Layer labels (right side)
\node[layerlabel, fill=layerstructural!15, text=layerstructural] at (7, -\lh) {Layer 0};
\node[layerlabel, fill=layerstructural!15, text=layerstructural!85] at (7, -2*\lh) {Layer 1};

\node[layerlabel, fill=layerstructural!15, text=layerstructural!70] at (7, -5*\lh) {Layer 2};
\node[layerlabel, fill=layerstructural!15, text=layerstructural!70] at (7, -9*\lh) {Layer 2};
\node[layerlabel, fill=layerstructural!15, text=layerstructural!70] at (7, -13*\lh) {Layer 2};

\node[layerlabel, fill=layercontent!15, text=layercontent] at (7, -6*\lh) {Layer 3};
\node[layerlabel, fill=layercontent!15, text=layercontent] at (7, -10*\lh) {Layer 3};
\node[layerlabel, fill=layercontent!15, text=layercontent] at (7, -14*\lh) {Layer 3};

\node[layerlabel, fill=layercontent!15, text=layercontent!80] at (7, -7*\lh) {Layer 4};
\node[layerlabel, fill=layercontent!15, text=layercontent!80] at (7, -11*\lh) {Layer 4};
\node[layerlabel, fill=layercontent!15, text=layercontent!80] at (7, -15*\lh) {Layer 4};

\node[layerlabel, fill=layercontent!15, text=layercontent] at (7, -16*\lh) {Layer 3};
\node[layerlabel, fill=layercontent!15, text=layercontent] at (7, -17*\lh) {Layer 3};

% Connecting lines
\draw[gray!40] (claude.west) -- ++(-0.15,0);
\draw[gray!40] (ctx1.west) -- ++(-0.15,0);
\draw[gray!40] (stages.west) -- ++(-0.15,0);
\draw[gray!40] (config.west) -- ++(-0.15,0);
\draw[gray!40] (shared.west) -- ++(-0.15,0);
\draw[gray!40] (setup.west) -- ++(-0.15,0);

\end{tikzpicture}
\caption{Folder structure of a typical ICM workspace, with layer annotations. Files and folders are color-coded by their role in the context hierarchy. Layer 3 material (reference) persists across runs. Layer 4 material (working artifacts) changes each time the pipeline executes.}
\label{fig:folders}
\end{figure}
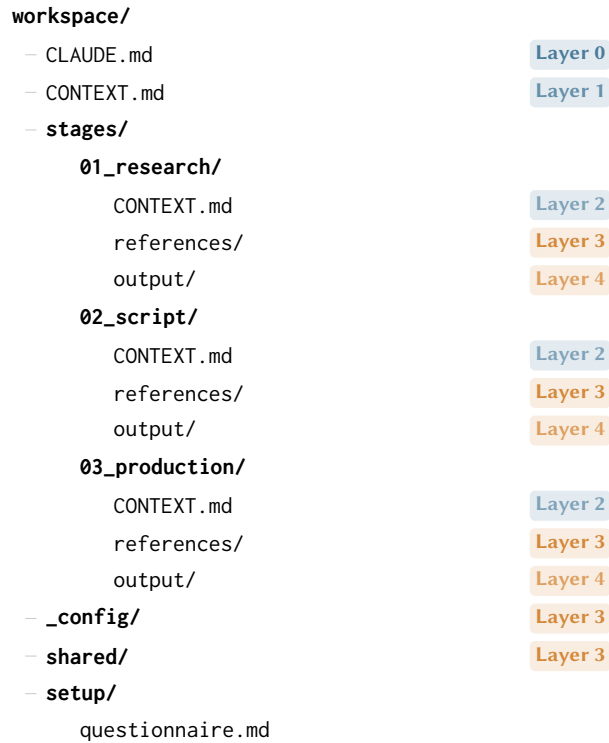

The folder structure for a typical workspace is shown in Figure~\ref{fig:folders}.

The numbering encodes execution order. The folder boundaries enforce separation of concerns. The \texttt{output/} directories are the Layer 4 handoff points: the output of stage 01 becomes available as input to stage 02. If a human edits a file in \texttt{01\_research/output/} before running stage 02, the agent picks up the edited version. The \texttt{references/} directories and \texttt{\_config/} folder hold Layer 3 material: the stable knowledge and constraints that persist across runs.

Layer 2 is the control point of the entire system. Each stage contract includes an Inputs table that specifies exactly which files from Layers 3 and 4 the agent should load, and which sections of those files are relevant.\footnote{Larger reference collections can include their own routing files, a CONTEXT.md within a configuration or design system folder, that help agents navigate to the right content within the collection. This is the routing pattern from Layer 1 applied recursively within Layer 3.} Without this scoping mechanism, an agent would either load everything in the workspace or rely on its own judgment about what matters. The Inputs table makes the selection explicit, editable, and auditable.

This is the filesystem doing the work that a framework would otherwise do in code. Stage sequencing is the folder numbering. Context scoping is the folder hierarchy. State management is the files on disk. Coordination between stages is one folder's output being another folder's input.

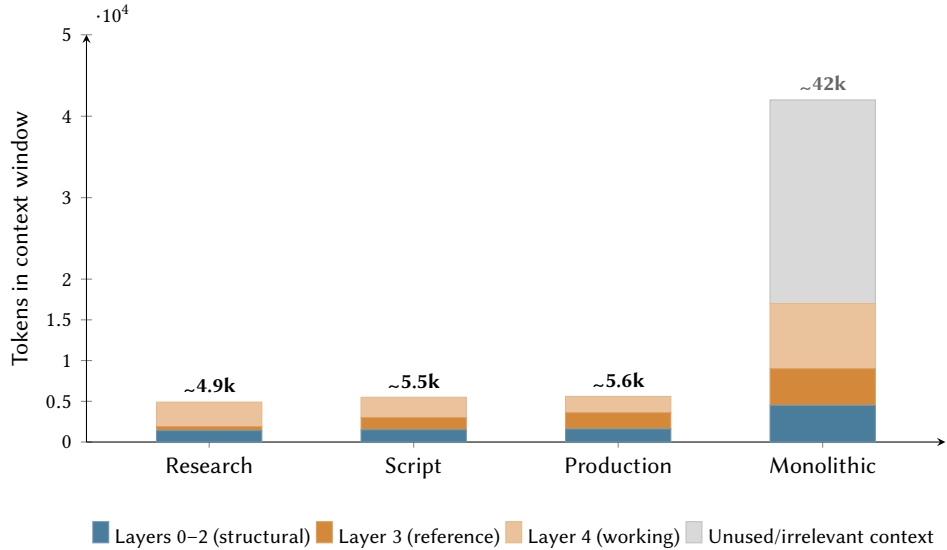
\begin{figure}[t]
\centering
\begin{tikzpicture}
\begin{axis}[
    ybar stacked,
    bar width=1.4cm,
    width=13cm,
    height=7cm,
    ylabel={Tokens in context window},
    ylabel style={font=\sffamily\small},
    symbolic x coords={Research, Script, Production, Monolithic},
    xtick=data,
    xticklabel style={font=\sffamily\small},
    ymin=0, ymax=50000,
    ytick={0, 5000, 10000, 15000, 20000, 30000, 40000, 50000},
    yticklabel style={font=\sffamily\footnotesize},
    legend style={
        at={(0.5,-0.18)},
        anchor=north,
        legend columns=4,
        font=\sffamily\footnotesize,
        draw=none
    },
    nodes near coords style={font=\sffamily\tiny, /pgf/number format/1000 sep={\,}},
    every node near coord/.append style={anchor=south, yshift=-2pt},
    axis lines=left,
    enlarge x limits=0.2,
    clip=false
]

% Layers 0-2 (structural)
\addplot[fill=layerstructural, draw=layerstructural!80] coordinates
    {(Research, 1400) (Script, 1500) (Production, 1600) (Monolithic, 4500)};

% Layer 3 (reference)
\addplot[fill=layercontent, draw=layercontent!80] coordinates
    {(Research, 500) (Script, 1500) (Production, 2000) (Monolithic, 4500)};

% Layer 4 (working)
\addplot[fill=layercontent!50, draw=layercontent!60] coordinates
    {(Research, 3000) (Script, 2500) (Production, 2000) (Monolithic, 8000)};

% All other context (monolithic overhead)
\addplot[fill=black!15, draw=black!25] coordinates
    {(Research, 0) (Script, 0) (Production, 0) (Monolithic, 25000)};

\legend{Layers 0--2 (structural), Layer 3 (reference), Layer 4 (working), Unused/irrelevant context}

% Total labels
\node[font=\sffamily\footnotesize\bfseries, above] at (axis cs:Research, 4900) {\texttildelow4.9k};
\node[font=\sffamily\footnotesize\bfseries, above] at (axis cs:Script, 5500) {\texttildelow5.5k};
\node[font=\sffamily\footnotesize\bfseries, above] at (axis cs:Production, 5600) {\texttildelow5.6k};
\node[font=\sffamily\footnotesize\bfseries, above, text=black!60] at (axis cs:Monolithic, 42000) {\texttildelow42k};

\end{axis}
\end{tikzpicture}
\caption{Context window composition by stage (representative token counts from the script-to-animation workspace). The three ICM stages each deliver 2,000--8,000 focused tokens. A monolithic approach loading all stages' instructions, all reference material, and all prior outputs produces a context window exceeding 40,000 tokens, most of it irrelevant to the current task.}
\label{fig:context}
\end{figure}

From the model's perspective, this layered loading changes the composition of the context window at each stage. Layers 0 through 2 together contribute roughly 1,300 to 1,600 tokens of identity, routing, and stage-specific instruction. Layer 3 adds reference material scoped to the current stage, typically 500 to 2,000 tokens depending on how many conventions and guidelines apply. Layer 4 adds the working material for this run, a research document, a script, a specification, which varies with the content but rarely exceeds a few thousand tokens when the previous stage has done its job of condensing and structuring. The total context delivered to the model at any given stage typically ranges from 2,000 to 8,000 tokens, well within the range where current models perform at their best. Figure~\ref{fig:context} illustrates this composition across three example stages and contrasts it with a monolithic approach.

Contrast this with a monolithic approach where all stage instructions, all reference files, and all prior outputs are loaded into a single prompt. That approach can easily reach 30,000 to 50,000 tokens, pushing into the range where Liu et al.\ found significant performance degradation on information retrieval tasks \cite{liu2024}. The ``unused/irrelevant context'' segment in the monolithic bar of Figure~\ref{fig:context} represents tokens from stages other than the one currently executing: instructions the agent will not follow during this step, reference material that applies to a different stage, and prior outputs already consumed by earlier stages. In ICM, these tokens are never loaded. In a monolithic prompt, they occupy context space without contributing to the current task. The compression research by Jiang et al.\ \cite{jiang2023, jiang2024} addresses this problem after the fact. ICM's architecture avoids it by construction: each stage receives a focused, appropriately sized context window because the folder structure determines what gets loaded.

Richard Gabriel argued that systems prioritizing simplicity of implementation over feature completeness tend to survive and spread, because they are easier to port, easier to understand, and easier to improve incrementally \cite{gabriel1991}. ICM trades the flexibility of a programmatic orchestrator for the portability, inspectability, and editability of plain files. That tradeoff is the point.

In the same spirit, Plan 9 from Bell Labs extended Unix's ``everything is a file'' principle to its full conclusion, representing all system resources as files in per-process namespaces \cite{pike1995}. ICM applies the same idea to AI workflows: all state, all context, all instructions exist as files in a folder namespace.

\subsection{Stage Contracts and Handoffs}

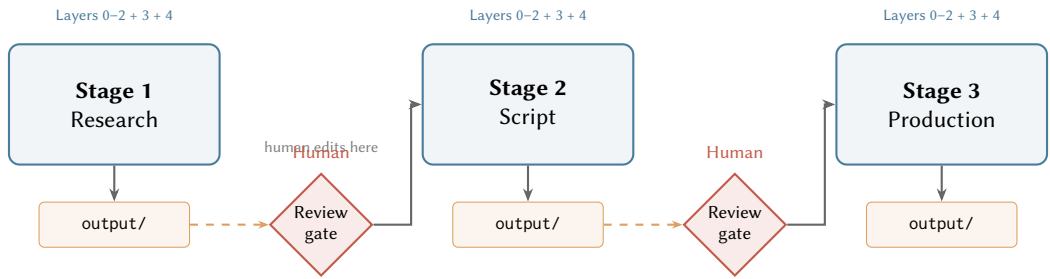
\begin{figure}[t]
\centering
\begin{tikzpicture}[
    stage/.style={
        draw=stagebox, fill=stagebox!8, thick,
        rounded corners=4pt, minimum width=2.8cm, minimum height=1.6cm,
        font=\sffamily\small, align=center
    },
    outputbox/.style={
        draw=layercontent!80, fill=layercontent!8,
        rounded corners=2pt, minimum width=2cm, minimum height=0.6cm,
        font=\ttfamily\scriptsize, align=center
    },
    review/.style={
        draw=reviewgate, fill=reviewgate!12, thick,
        diamond, minimum width=1cm, minimum height=1cm,
        font=\sffamily\scriptsize, align=center, inner sep=1pt
    },
    arrow/.style={-{Stealth[length=5pt]}, thick, color=black!60},
    dataarrow/.style={-{Stealth[length=4pt]}, thick, color=layercontent!80, dashed}
]

% Stages
\node[stage] (s1) at (0,0) {\textbf{Stage 1}\\Research};
\node[stage] (s2) at (5.5,0) {\textbf{Stage 2}\\Script};
\node[stage] (s3) at (11,0) {\textbf{Stage 3}\\Production};

% Output folders
\node[outputbox] (o1) at (0,-1.6) {output/};
\node[outputbox] (o2) at (5.5,-1.6) {output/};
\node[outputbox] (o3) at (11,-1.6) {output/};

% Review gates
\node[review] (r1) at (2.75,-1.6) {Review\\gate};
\node[review] (r2) at (8.25,-1.6) {Review\\gate};

% Arrows: stage to output
\draw[arrow] (s1) -- (o1);
\draw[arrow] (s2) -- (o2);
\draw[arrow] (s3) -- (o3);

% Arrows: output through review to next stage
\draw[dataarrow] (o1) -- (r1);
\draw[arrow] (r1.east) -- ++(0.5, 0) |- (s2.west);
\draw[dataarrow] (o2) -- (r2);
\draw[arrow] (r2.east) -- ++(0.5, 0) |- (s3.west);

% Edit label
\node[font=\sffamily\tiny, text=black!50, above=1pt] at (2.75, -0.8) {human edits here};

% Human icons (simple)
\node[font=\sffamily\scriptsize, text=reviewgate, above=2pt of r1] {\textsf{Human}};
\node[font=\sffamily\scriptsize, text=reviewgate, above=2pt of r2] {\textsf{Human}};

% Layer labels above stages
\node[font=\sffamily\tiny, text=layerstructural, above=4pt of s1] {Layers 0--2 + 3 + 4};
\node[font=\sffamily\tiny, text=layerstructural, above=4pt of s2] {Layers 0--2 + 3 + 4};
\node[font=\sffamily\tiny, text=layerstructural, above=4pt of s3] {Layers 0--2 + 3 + 4};

\end{tikzpicture}
\caption{Pipeline flow through three stages with review gates. Each stage receives its own context (Layers 0--4), writes output to its folder, and the human reviews and optionally edits before the next stage reads it. The same model executes every stage; the folder structure controls what context it receives.}
\label{fig:pipeline}
\end{figure}

Figure~\ref{fig:pipeline} illustrates the flow between stages. Each stage reads from the previous stage's output folder, processes it according to its own contract, and writes to its own output folder. At each boundary, the human can inspect and edit the output before the next stage runs.

Each stage in an ICM workspace defines a contract with three parts: what it reads (inputs), what it does (process), and what it writes (outputs). This contract is spelled out in the stage's \texttt{CONTEXT.md} file.

A typical stage contract looks like this:

\begin{lstlisting}
## Inputs
- Layer 4 (working): ../01_research/output/
- Layer 3 (reference): ../../_config/voice.md
- Layer 3 (reference): references/structure.md

## Process
Write a script based on the research output.
Follow the structure in structure.md.
Match the tone described in voice.md.

## Outputs
- script_draft.md -> output/
\end{lstlisting}

The Inputs table distinguishes between Layer 3 files (reference material that stays the same every run) and Layer 4 files (working artifacts from this specific run). The agent reads the \texttt{CONTEXT.md}, follows the instructions, and writes its output. The human reviews what landed in \texttt{output/}. If it needs adjustment, the human edits the file directly. The next stage reads whatever is there.

This implements prompt chaining at the filesystem level. Wu, Terry, and Cai introduced AI Chains as a method for creating transparent, controllable multi-step LLM workflows where each step's output becomes the next step's input \cite{wu2022chains}. ICM does the same thing, but the chain is a sequence of folders and the links between them are plain files. The stage outputs serve as intermediate representations: each one is a complete, readable artifact that captures the work done so far and provides everything the next stage needs to continue.

There is also something of Knuth's literate programming in this design \cite{knuth1984}. The markdown files that instruct the agent are simultaneously the documentation that tells a human what the stage does, what it expects, and what it produces. The instruction set and the documentation are the same artifact. This is useful in practice because it means the workspace is self-documenting. A new team member can read the \texttt{CONTEXT.md} files top to bottom and understand the entire pipeline without running it.

Wei et al. demonstrated that breaking complex reasoning into intermediate steps dramatically improves LLM performance \cite{wei2022}. ICM applies this finding architecturally: complex workflows are decomposed into stages with explicit boundaries, and each stage receives focused, stage-appropriate context. The model gets a clear, scoped task at each step rather than a monolithic instruction to do everything in a single pass.

\subsection{Portability and Reproducibility}

A workspace is a folder. It can be copied to another machine, committed to Git, emailed as a zip file, or synced through any cloud storage service. It carries its own prompts, its own context structure, its own stage definitions. There is no server to configure, no environment to replicate, no deployment step.

ICM workspaces are Git-compatible by default \cite{chacon2014}. Every change to a prompt, every edit to a stage output, every configuration adjustment is diffable and reversible. Stage outputs can be committed after each run, creating a version history of the entire production pipeline's behavior over time. This is infrastructure as code \cite{morris2021} applied to AI workflows: the workspace definition is the system. There is no separate deployment artifact.

This portability matters for a practical reason. If a consultant builds a workspace for a client's weekly reporting workflow, handing it over means copying a folder. The client can run it, edit the prompts to match their evolving needs, and adjust stages without involving a developer. The same handoff with a framework-based solution typically requires documentation, environment setup, dependency management, and ongoing technical support.

\section{Working Implementations}

ICM is not a theoretical proposal. The protocol has been implemented and tested across several production workflows.\footnote{All workspaces referenced here are available or buildable through the ICM repository at \url{https://github.com/RinDig/Interpretable-Context-Methodology-ICM-}.}

\subsection{Model and Environment}

All workspaces described here were developed and run using Claude Code with Claude Opus 4.6 as the primary agent \cite{anthropic2026opus}. For sub-agent tasks within stages, Opus 4.6 delegates to Claude Sonnet 4.6 through its Agent Teams capability, which coordinates multiple agents working in parallel from a single orchestrator.

A detail worth noting: Opus 4.6 uses the workspace's own context files, the \texttt{CONTEXT.md} hierarchy and Layer 3 reference material, to fill prompts for its sub-agents. The model reads the folder structure to determine what context each sub-agent should receive and what task it should perform. This means the ICM architecture is doing double duty. It structures context for the primary agent, and it provides the specification that the primary agent uses to delegate work. The folder hierarchy is both the human's control surface and the model's orchestration logic.

ICM is designed to be model-agnostic. The protocol specifies folder structure, file formats, and naming conventions. It does not depend on any model-specific capability. A workspace built for Claude could be run with a different model by pointing that model at the same files. Whether the results would be equivalent is an empirical question that depends on how different models handle the same context, but the protocol itself imposes no vendor lock-in. The workspaces described below were tested with the models listed above.

\subsection{Script-to-Animation Pipeline}

The first workspace built on ICM takes a content idea through three stages to produce a working animated video.

Stage 1 (\texttt{01\_research}) takes a topic and produces structured research output: key points, narrative angles, supporting data. The agent reads a research brief from the user and writes a research document to its output folder.

Stage 2 (\texttt{02\_script}) reads the research output and writes a script. The stage's \texttt{CONTEXT.md} points the agent to a voice guide and structural template in the \texttt{\_config/} folder. The script follows the user's established tone and format.

Stage 3 (\texttt{03\_production}) reads the finished script and produces animation specifications and working Remotion\footnote{Remotion is a React-based framework for creating videos programmatically.} code. The stage's context includes design guidelines, color palettes, and animation conventions from setup.

At each stage boundary, the human reviews the output. A research document that misses an important angle gets edited before the script stage runs. A script that runs too long gets trimmed before the production stage sees it. The agent at each stage works with whatever the human left in the previous output folder.

This workspace runs on a single Claude Code session. One orchestrating agent (Opus 4.6) manages the pipeline, delegating sub-tasks within stages to faster sub-agents (Sonnet 4.6) as described in Section 4.1. The delegation is itself driven by the folder structure: the orchestrating agent reads the stage's \texttt{CONTEXT.md} to determine what work to delegate and what context to provide. There is no separate orchestration framework. The same folder hierarchy that tells the human what each stage does tells the agent how to coordinate its sub-agents. In compiler terms, the workspace performs multi-pass compilation: the processing engine runs multiple times, producing a different intermediate representation at each pass, with the folder structure determining which pass runs next.

\subsection{Course Deck Production}

A second workspace takes unstructured source material (PDFs, papers, lecture notes, rough outlines) and produces polished PowerPoint slide decks through five stages: content extraction, structural planning, slide drafting, visual design specification, and final assembly.

The five-stage structure matters because slide deck production is a process where human judgment is essential at several points. The structural plan (stage 2 output) determines the entire arc of the presentation. Getting it wrong means everything downstream is wrong. By surfacing the structural plan as an editable markdown file before any slides are drafted, ICM lets the human course-correct at the point where correction is cheapest and most effective.

\subsection{Building New Workspaces}

ICM includes a workspace-builder: a five-stage workspace whose output is a new workspace. It walks through discovery (what is the domain, what is the workflow), stage mapping (where are the natural breakpoints), scaffolding (creating the folder structure), questionnaire design (what setup questions should the workspace ask), and validation (does the pipeline run end to end).

The workspace-builder itself follows ICM conventions. The workspaces it produces are consistent because the builder enforces the same structural rules it was built with.

This means practitioners can create new workspaces for their own domains without understanding the underlying conventions in detail. The builder encodes the conventions into its process. A marketing team can build a workspace for campaign production. A research group can build one for literature review and synthesis. A consultancy can build one for client deliverable pipelines. Each workspace is a folder they own and control.

ICM workspaces have been adopted by groups outside the author's organization. Researchers at the University of Edinburgh's Neuropolitics Lab have built workspaces for their domain, and teams at ICR Research and the Academy of International Affairs in Bonn are developing workspaces for their own workflows. The details of these implementations are limited by nondisclosure agreements, but their existence is noted here because the reviewer's natural question, does ICM work when someone other than its designer builds and operates the workspace, has at least a preliminary answer: yes, across academic research, policy analysis, and content production. A structured study of these external deployments is a clear next step.

\subsection{Early Practitioner Experience}

ICM has been used in production across content creation, training material development, research analysis, and policy workflows. The observations reported here are drawn from an invite-only practitioner community of 52 members whose backgrounds range from AI engineers and software developers to business owners, content creators, and academic researchers. These observations come from ongoing conversations with community members rather than from formal data collection protocols. They should be read as practitioner reports rather than controlled findings, but they reflect a broader base of experience than the author's own use alone.

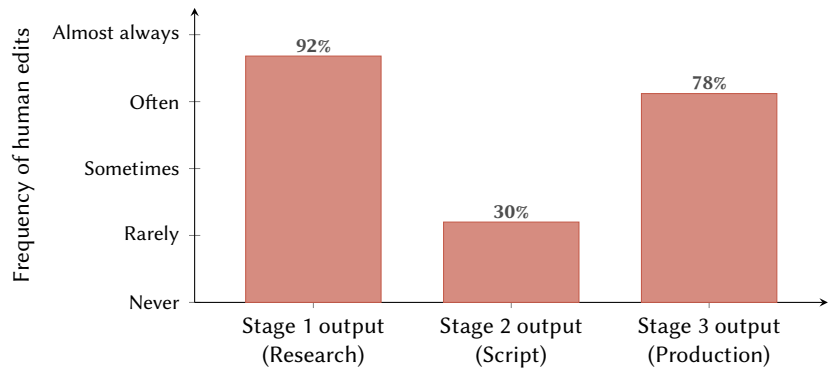
\begin{figure}[t]
\centering
\begin{tikzpicture}
\begin{axis}[
    ybar,
    bar width=1.8cm,
    width=10cm,
    height=5.5cm,
    ylabel={Frequency of human edits},
    ylabel style={font=\sffamily\small},
    symbolic x coords={Stage 1 output\\(Research), Stage 2 output\\(Script), Stage 3 output\\(Production)},
    xtick=data,
    xticklabel style={font=\sffamily\small, align=center},
    ymin=0, ymax=110,
    ytick={0, 25, 50, 75, 100},
    yticklabels={Never, Rarely, Sometimes, Often, Almost always},
    yticklabel style={font=\sffamily\footnotesize},
    axis lines=left,
    enlarge x limits=0.3,
    nodes near coords,
    nodes near coords style={font=\sffamily\footnotesize\bfseries, above, yshift=-2pt},
    every node near coord/.append style={text=black!70},
    point meta=explicit symbolic,
]
\addplot[fill=reviewgate!70, draw=reviewgate] coordinates
    {(Stage 1 output\\(Research), 92) [92\%]
     (Stage 2 output\\(Script), 30) [30\%]
     (Stage 3 output\\(Production), 78) [78\%]};
\end{axis}
\end{tikzpicture}
\caption{Observed frequency of human edits at each stage boundary, reported by 33 practitioners using multi-stage ICM workspaces. Intervention follows a U-shaped pattern: heavy at stage 1 (direction-setting), light at middle stages (constrained execution), heavy again at the final stage (aligning output with earlier decisions). Stage 1 editing is creative judgment. Final-stage editing is closer to debugging. Values are approximate and based on practitioner self-report through conversation, not instrumented measurement.}
\label{fig:intervention}
\end{figure}

The most consistent observation is where people choose to intervene (Figure~\ref{fig:intervention}). Across 33 community members who have used the script-to-animation workspace or structurally similar multi-stage workspaces, 30 report an intervention pattern consistent with a U-shape: heavy editing at stage 1 (direction-setting), light editing at the middle stages, and heavy editing again at the final stage (aligning output with earlier decisions). The remaining three report roughly equal editing across all stages. These numbers come from practitioner conversations, not from instrumented measurement, and should be interpreted accordingly.

The two peaks reflect different kinds of editing. Stage 1 editing is directional: the user is narrowing from broad possibilities to a specific angle, deciding what the piece is about. This is creative judgment. Final-stage editing is alignment work: the user is checking that the output faithfully represents decisions made in earlier stages. This is closer to debugging. The practitioner traces a misalignment in the output back through the pipeline to find where it diverged from the source material. Section 6 explores what tooling for this kind of traceability might look like.

The middle stages get the lightest touch because they sit between well-defined anchors. The earlier stage output sets the direction. The reference material (Layer 3 voice guides, structural templates) constrains the execution. With both anchors in place, the middle stages have less room to go wrong, and practitioners tend to trust them. This aligns with Parasuraman, Sheridan, and Wickens's observation that appropriate automation levels vary by task function \cite{parasuraman2000}.

A second pattern involves prompt editing. Non-technical users, people without development experience, have successfully modified stage behavior by editing the markdown \texttt{CONTEXT.md} files. Changes include adjusting tone instructions, adding constraints (``keep scripts under 90 seconds''), and reordering the emphasis within a stage's process description. These edits would be equivalent to modifying agent configuration in a framework-based system, a task that typically requires a developer. The plain-text interface lowers this barrier in practice.

A third pattern is worth noting for its implications about accessibility. Three community members with no prior coding experience and no previous exposure to Claude Code used the ICM workspace-builder's questionnaire and setup process to create and run workspaces that produced ten-minute animated videos from scripts. They edited \texttt{CONTEXT.md} files, reviewed stage outputs, and iterated on their workspaces without developer assistance. This is a single data point from a small group, but it suggests that the filesystem interface can make AI agent orchestration accessible to people who would not be able to use a framework-based system at all.

A fourth pattern is workspace duplication. Users who have a working workspace for one content format (say, short explainer videos) duplicate the folder, modify the stage prompts to target a different format (say, long-form essays), and run the new workspace without rebuilding from scratch. The workspace-builder supports creating workspaces from nothing, but in practice people often prefer to copy and adapt an existing one. This mirrors how Unix users build new shell scripts by modifying existing ones rather than starting from a blank file.

These observations are drawn from a community of varied backgrounds across a growing but still limited set of workflow types. A structured evaluation with formal data collection, systematic interviews, and controlled comparisons would be needed to draw firm conclusions about the generality of these patterns. The observations are reported here because they informed the protocol's design evolution and because they suggest directions for future study.

\subsection{Threats to Validity}

Several limitations constrain the conclusions that can be drawn from the current work, and naming them is important for interpreting the results above. The practitioner community provides a broader evidence base than single-author observation, but data collection has been informal: observations come from ongoing conversations rather than structured interviews, diary studies, or instrumented usage logging. The community is invite-only and self-selected, introducing both selection bias and potential enthusiasm bias. The reported intervention patterns (30 of 33 practitioners observing a U-shape) are self-reported through conversation and have not been verified through controlled measurement.

While ICM has been adopted across content production, academic research, and policy analysis workflows, the majority of active use remains concentrated in content production. The academic and policy deployments are early-stage, and their outcomes cannot yet be reported in detail. All testing was conducted using a single model family (Claude Opus 4.6 and Sonnet 4.6). Cross-model evaluation is a natural next step but falls outside the scope of this paper, which focuses on the architectural pattern and its interaction properties rather than model-specific performance. Output quality may vary with other models, particularly those with different context-handling characteristics.

No controlled comparison has been conducted between ICM's staged context loading and a monolithic prompting approach on the same tasks, so the claim that scoped context improves output quality rests on the theoretical support from the ``lost in the middle'' literature \cite{liu2024} and practitioner judgment rather than measured effect sizes. A formal user study with systematic data collection, structured interviews, and controlled comparisons across varied workflow types and participant backgrounds would substantially strengthen the empirical foundations of this work.

\section{Discussion}

\subsection{Where This Works}

ICM handles sequential multi-step workflows where a human reviews output at each stage. In practice, the protocol has been applied to content production pipelines (script-to-animation, short-form video), training material development (slide deck generation from source material), academic research workflows (at the University of Edinburgh and ICR Research), and policy analysis (at the Academy of International Affairs, Bonn). The common thread across these deployments is that the workflows are sequential, the outputs benefit from human review at each step, and the same pipeline runs repeatedly with different input.

The common thread is that these workflows are sequential (step 2 follows step 1), reviewable (a human should check each step's output), and repeatable (the same pipeline runs weekly or daily with different input). For this class of workflow, ICM provides full orchestration capability with no framework code, no server infrastructure, and no developer dependency for day-to-day operation.

\subsection{Where This Does Not Work}

ICM is not a replacement for multi-agent frameworks in every context.

Real-time multi-agent collaboration, where agents need to communicate dynamically and respond to each other's outputs in tight loops, requires the kind of message-passing infrastructure that AutoGen \cite{wu2023autogen} and similar frameworks provide. ICM's sequential, file-based handoffs are too slow for this.

High-concurrency systems where many users hit the same pipeline simultaneously need proper queueing, state isolation, and deployment infrastructure. ICM is local-first by design. Scaling it to concurrent users would require building the infrastructure ICM was designed to avoid.

Workflows that require complex branching logic based on AI decisions mid-pipeline are awkward in ICM. A human can make branching decisions between stages (run stage 3a instead of 3b based on what they see in the stage 2 output), but automated branching would require scripting that moves ICM toward being a framework itself.

These boundaries matter. The claim is not that ICM replaces existing tools across the board. The claim is that for a large and common class of workflows, the existing tools provide more complexity than the problem requires, and that complexity has real costs: opacity, fragility, developer dependency, and overhead that slows iteration.

\subsection{Observability as a Side Effect}

The most useful property of ICM may be one that was not designed as a feature. Because every intermediate output is a plain file, the system is observable by default. There is no logging layer to build, no dashboard to configure, no special tooling to inspect pipeline state. You open a folder and read the files.

Rudin argued that inherently interpretable systems should be preferred over post-hoc explanations of opaque ones \cite{rudin2019}. ICM is a glass-box AI workflow. It did not become transparent through the addition of an explanation layer. It was never opaque in the first place, because every artifact is a plain-text file that a human can read.

Amershi et al.'s guidelines for human-AI interaction include ``make clear what the system can do,'' ``support efficient correction,'' and ``support efficient dismissal'' \cite{amershi2019}. Stage contracts make capabilities explicit. Markdown files support efficient correction (open, edit, save). Review gates at every stage boundary support dismissal (decide not to proceed, re-run the previous stage with different input, or abandon the run entirely).

The regulatory landscape may also be relevant here. The EU AI Act's human oversight requirements \cite{enqvist2023, novelli2024} emphasize staged review, audit trails, and defined intervention points. ICM produces these as a byproduct of its architecture: there is no way to run an ICM pipeline without generating inspectable intermediate artifacts, because the intermediate artifacts are how the stages communicate. Whether this constitutes compliance with specific regulatory requirements is a legal question this paper does not attempt to answer, but the structural alignment is worth noting.

\subsection{Implications for Intelligent System Design}

The discussion so far has focused on how ICM structures the human side of human-AI interaction: edit surfaces, review gates, observability. But the architecture also has implications for how the intelligent system itself performs, and these are worth examining.

The core mechanism is context scoping. By delivering different context to the same model at each stage, ICM changes the task the model is performing. A model that receives research instructions, source material, and a topic brief behaves differently from the same model receiving a script template, a voice guide, and a research summary. The model's capabilities do not change between stages. What changes is the information it has available when generating output. This is context engineering in practice: the performance of the system depends on what context is delivered, in what structure, and at what moment.

The Layer 3/Layer 4 distinction adds a further dimension. Reference material (Layer 3) and working artifacts (Layer 4) ask different things of the model. Reference material says: here are the rules, follow them. Working artifacts say: here is the input, transform it. Delivering these as structurally separate context, rather than mixing them in a single undifferentiated prompt, gives the model clearer signals about which information constrains its behavior and which information it should act on. Whether this structural separation measurably improves output quality compared to a flat context of equivalent content is an open empirical question, but early practitioner experience suggests that stages where reference and working material are clearly separated produce more consistent adherence to style and format guidelines.

This raises a question about the relationship between context structure and output quality. In early use, a pattern emerged: stages with tightly scoped context (clear instructions, limited reference material, a specific output format) produced more consistent results than stages with broad context (open-ended instructions, large volumes of reference material, loosely defined output expectations). This is consistent with the ``lost in the middle'' findings \cite{liu2024} and with the chain-of-thought literature showing that decomposed tasks outperform monolithic ones \cite{wei2022}, but it suggests something more specific. The structure of the context delivery, how information is organized and bounded, may matter as much as the content of the context itself. ICM's folder-based scoping enforces this structure by default: each stage folder contains only what that stage needs, and the boundaries are visible and editable.

There are open questions here that the current work does not answer. First, does the five-layer hierarchy (workspace identity, task routing, stage contracts, reference material, working artifacts) generalize across model families, or is it tuned to the specific attention patterns of the models tested? The protocol is designed to be model-agnostic (Section 4.1), but all current testing has been conducted on a single model family. Cross-model evaluation, running the same workspace on Claude, GPT, Gemini, and open-weight models such as Llama, is a clear next step. This paper scopes that question as future work because the present contribution is the architectural pattern and its interaction properties, not a model-specific performance claim. Second, as context windows grow larger, does selective loading become less important? If a model can reliably attend to 200,000 tokens without degradation, the engineering argument for ICM's scoping weakens, though the human-interaction arguments (observability, editability, review gates) remain. Third, how sensitive is stage output quality to the ordering and formatting of context within a layer? The current protocol specifies what files a stage should load but does not prescribe the order in which they appear in the context window. Whether ordering matters at the scale of ICM's typical context sizes (2,000 to 8,000 tokens per stage) is an empirical question worth investigating.

These questions point toward a research program that sits at the intersection of context engineering and interaction design: understanding how the structure of information delivery to language models affects both the model's output quality and the human's ability to steer, inspect, and correct that output. ICM provides a concrete platform for investigating these questions because its architecture makes the context structure explicit, editable, and observable at every stage.

\section{Future Directions: Compilation, Debugging, and Source Integrity}

The previous sections describe ICM as it currently works in production. This section describes where it should go next. The ideas here are informed by early practitioner experience and by a structural analogy that the paper has not yet drawn: the relationship between ICM workspaces and multi-pass compilers.

\subsection{ICM as Multi-Pass Incremental Compilation}

The paper has grounded ICM in Unix pipelines, Make, and the pipe-and-filter pattern. There is a closer analogy that deserves attention: multi-pass compilation \cite{aho2006}.

A multi-pass compiler transforms source code through a sequence of discrete passes. The lexer produces tokens. The parser produces a syntax tree. Semantic analysis annotates the tree. Optimization passes rewrite it. Code generation produces the final output. Each pass reads the output of the previous pass, transforms it according to its own rules, and writes an intermediate representation that the next pass can consume. The intermediate representations are well-defined, inspectable, and (in debugging builds) preserved for examination.

ICM does the same thing with content. Stage 1 (research) transforms a topic brief into structured research output. Stage 2 (script) transforms the research output into a script. Stage 3 (production) transforms the script into animation specifications and code. Each stage reads the previous stage's output, applies its own context and instructions, and writes an intermediate artifact that the next stage consumes. The intermediate artifacts are plain files that can be opened, read, and edited.

The analogy extends further. Incremental compilation means recompiling only the parts of the program that changed, rather than rebuilding from scratch. ICM supports this by default: if the research output is fine but the script needs rework, the practitioner re-runs stage 2 without touching stage 1. If a voice guide in the reference material changes, only the stages that load that file need to run again. The folder structure tracks these dependencies implicitly: a stage's Inputs table declares which files it reads, and a change to any of those files signals that the stage's output may be stale.

This is worth naming because it connects ICM to a body of compiler engineering that has spent fifty years solving the problems of pass decomposition, intermediate representation design, and selective recompilation. The current paper draws from Unix and software architecture. Future work should draw from compiler theory as well, particularly around dependency tracking, change propagation, and the formal properties of intermediate representations.

\subsection{Toward Semantic Debugging}

Traditional debugging rests on a simple principle: when the output is wrong, trace the failure back through the program's execution to find the instruction that caused it \cite{zeller2009}. Debuggers provide tools for this: breakpoints that pause execution at specific instructions, stack traces that show the call chain, variable inspection that shows state at each point, and source maps that connect compiled output back to the original source.

ICM currently provides observability but not traceability. A practitioner can open any stage's output folder and read what the agent produced. But if a phrase in the stage 3 output sounds wrong, there is no direct way to trace that phrase back to the specific instruction, reference file, or previous stage output that caused it. The practitioner has to do this manually: read the stage 3 contract, check which files it loaded, read those files, and form a judgment about which source is responsible. This works. It is how most ICM debugging happens today. But it is the equivalent of debugging a program by reading the source code and thinking hard, without a debugger.

The question is what a debugger for semantic content would look like. Several directions are worth exploring.

\textbf{Output provenance through identifiers.} If each section of a stage's output carried an identifier linking it to the source instruction or reference file that produced it, a practitioner could trace backward from any part of the output to the context that generated it. In compiler terms, this is the equivalent of debug symbols or source maps: metadata that connects output back to source without changing the output itself. In practice, this could mean embedding lightweight markers (GUIDs, section tags, or comment annotations) in stage output files that reference specific sections of the stage's \texttt{CONTEXT.md} or Layer 3 reference files.

\textbf{Cross-stage trace verification.} In the script-to-animation workspace described in Section 4, one recurring problem has been misalignment between the animation specification (stage 3 output) and the script (stage 2 output). Timing drifts. Animations reference phrases that were revised. Visual density does not match pacing. This is the source of the U-shaped intervention pattern observed in Section 4.4: the stage 3 editing that brings intervention back up is almost entirely alignment work, tracing the final output back through the pipeline to find where it diverged from earlier decisions. The current solution is an audit file that forces the agent to trace back from the specification to the original script, re-verifying timing for each phrase and flagging inconsistencies. This works well enough that it catches most alignment errors, and the errors it catches are remarkably consistent in kind: frame count discrepancies, visual density mismatches, and pacing breaks at scene boundaries.

This audit file is a proto-debugger. It implements a specific kind of cross-stage verification: checking that the output of stage $n$ is consistent with the output of stage $n-2$ by re-reading both and comparing them against defined criteria. The pattern could be generalized. A stage contract could include a \texttt{Verify} section alongside its \texttt{Inputs}, \texttt{Process}, and \texttt{Outputs} sections, specifying which earlier stage outputs should be checked for consistency and what criteria to check against. The agent would run these verification checks as part of the stage's execution and flag discrepancies before the human reviews.

\textbf{Breakpoints in markdown.} The most speculative direction involves something like breakpoints for markdown files. In a traditional debugger, a breakpoint says ``pause here and let me inspect the state.'' In ICM, a breakpoint in a \texttt{CONTEXT.md} file might say ``after the agent processes this instruction, show me what it produced before continuing.'' This would be particularly useful in stages with complex instructions where the practitioner wants to verify that the agent interpreted a specific constraint correctly before it finishes the rest of the stage's work. It turns a single-pass stage into a sequence of verifiable sub-steps.

These ideas are not yet implemented. They are described here because the gap they address, the ability to trace output back to source, is a gap that compiler engineering solved decades ago and that ICM will need to solve as workspaces grow more complex.

\subsection{Source Integrity and the Edit-Source Principle}

The current paper describes ICM's review gates as places where practitioners edit stage output. This is useful and it works. But there is an argument, drawn directly from software engineering practice, that the source files should be what improves over time, and that editing output is treating symptoms rather than causes.

The argument is straightforward. If a script sounds wrong at stage 2, there are two possible responses. The first is to edit the script directly: fix the tone, adjust the phrasing, move on. The second is to ask why the script sounds wrong and trace the problem back to the source that produced it. Maybe the voice guide in the reference material is underspecified. Maybe the stage contract's instructions emphasize the wrong quality. Maybe the research output from stage 1 framed the topic in a way that led the script in the wrong direction. Editing the output fixes this run. Editing the source fixes every future run.

In compiler terms, editing the output is patching the binary. It works, but it does not improve the compiler. A developer who finds a bug in compiled code traces it back to the source and fixes it there, so that every subsequent build is correct.

For ICM, the tension is real. Creative content is fuzzier than compiled code. Sometimes the output needs a human touch that cannot be reduced to a source-level rule. A script might benefit from a turn of phrase that no amount of voice guide refinement would have produced. Editing the output in that case is the right move. The practitioner is adding value that the system cannot generate on its own.

But there is a class of output edits that are diagnostic. If the practitioner consistently tightens the opening paragraph, that is a signal that the stage contract should say ``keep the opening under three sentences.'' If the tone drifts formal every time, that is a signal that the voice guide needs a stronger example of the target register. These recurring edits are debugging information. They point to fixable source-level problems.

A future version of ICM could support this by tracking output edits across runs. If a practitioner edits the same kind of thing in the same stage's output three runs in a row, the system could surface that pattern and suggest a source-level change: a contract amendment, a reference file update, a new constraint. This would close the loop between output editing and source improvement, turning one-off fixes into durable system improvements.

The principle matters because it addresses a question about ICM's long-term trajectory. If workspaces are only as good as the last human edit of their output, they remain tools. If workspaces improve their own source files over time, incorporating the patterns they learn from human corrections, they become systems that get better with use. The debugging and traceability infrastructure described in the previous subsection is a prerequisite for this: you cannot improve the source if you cannot trace the problem back to it.

\section{Conclusion}

The principles that made Unix pipelines effective in the 1970s apply to AI agent orchestration in the 2020s. Programs that do one thing. Output of one becomes input of another. Plain text as universal interface. Human-readable intermediate state.

ICM applies these principles to a specific problem: structuring context for AI agents across multi-step workflows. The result is a system where the folder structure replaces the framework. One agent reads different context at each stage rather than multiple agents coordinating through code. Local scripts handle the mechanical work that does not need AI. Every intermediate output is a file a human can read and edit.

For practitioners whose AI workflows are sequential, reviewable, and repeatable, this means full pipeline capability with no framework to learn, no server to maintain, and no developer needed for day-to-day operation. The workspace is a folder. It can be copied, versioned, shared, and edited with a text editor. The simplest viable architecture for this class of problem is one that already exists on every computer: the filesystem.

The protocol is open source under the MIT license and includes a workspace-builder for creating new workspaces across any domain.

\end{document}